\documentclass[10pt,twocolumn,letterpaper]{article}

\usepackage[dvipsnames, svgnames, x11names]{xcolor}
\usepackage[final]{iccv}

\usepackage[normalem]{ulem}

\usepackage{graphicx}
\usepackage{amsmath}
\usepackage{amssymb}
\usepackage{booktabs}
\usepackage{marvosym}
\usepackage{colortbl}
\usepackage{amsxtra}

\definecolor{cvprblue}{rgb}{0.21,0.49,0.74}
\usepackage[pagebackref,breaklinks,colorlinks,allcolors=cvprblue]{hyperref}

\def\paperID{XXXX} 
\def\confName{ICCV}
\def\confYear{2025}

\begin{document}

\title{UrbanCraft: Urban View Extrapolation via Hierarchical Sem-Geometric Priors}
\vspace{-2ex}
\author{Tianhang Wang\textsuperscript{1}, Fan Lu\textsuperscript{1}, Sanqing Qu\textsuperscript{1}, Guo Yu\textsuperscript{1}, Shihang Du\textsuperscript{1}, Ya Wu\textsuperscript{2}, Yuan Huang\textsuperscript{2}, Guang Chen\textsuperscript{1,\Letter}\thanks{Corresponding author: guangchen@tongji.edu.cn}\\
\textsuperscript{1}Tongji University, \textsuperscript{2}China National Nuclear Corporation  \\
}

\maketitle

\begin{abstract}

Existing neural rendering-based urban scene reconstruction methods mainly focus on the Interpolated View Synthesis (IVS) setting that synthesizes from views close to training camera trajectory. However, IVS can not guarantee the on-par performance of the novel view outside the training camera distribution (\textit{e.g.}, looking left, right, or downwards), which limits the generalizability of the urban reconstruction application. Previous methods have optimized it via image diffusion, but they fail to handle text-ambiguous or large unseen view angles due to coarse-grained control of text-only diffusion. In this paper, we design UrbanCraft, which surmounts the Extrapolated View Synthesis (EVS) problem using hierarchical sem-geometric representations serving as additional priors. Specifically, we leverage the partially observable scene to reconstruct coarse semantic and geometric primitives, establishing a coarse scene-level prior through an occupancy grid as the base representation. Additionally, we incorporate fine instance-level priors from 3D bounding boxes to enhance object-level details and spatial relationships. Building on this, we propose the \textbf{H}ierarchical \textbf{S}emantic-Geometric-\textbf{G}uided Variational Score Distillation (HSG-VSD), which integrates semantic and geometric constraints from pretrained UrbanCraft2D into the score distillation sampling process, forcing the distribution to be consistent with the observable scene. Qualitative and quantitative comparisons demonstrate the effectiveness of our methods on EVS problem.

\end{abstract}

\section{Introduction}

Recent advances in neural implicit representations and rendering techniques, such as NeRF~\cite{mildenhall2021nerf}, have facilitated precise, high-quality 3D scene reconstruction and novel view synthesis. However, these methods assume specific conditions, including static scenes and dense, comprehensive sets of training images for precise scene reconstruction. 
To alleviate the stringent training set requirements, numerous strategies have been proposed to train NeRFs using a limited number of sparsely distributed images~\cite{yu2021pixelnerf, jain2021putting, niemeyer2022regnerf, yang2023freenerf, wang2023sparsenerf}. Nonetheless, these methods primarily concentrate on the limited quantity of training cameras instead of their pose distribution, which may likewise raise pose issues when skewed towards a specific location or perspective.

Recently, several methods~\cite{tancik2022block, turki2022mega, zhou2024drivinggaussian} have introduced specialized solutions for urban scene reconstruction, leveraging NeRF or 3DGS-based techniques. Most of these approaches focus on either reconstructing scenes with dynamic objects~\cite{xie2023s, yang2023emernerf, turki2023suds} or improving the efficiency of neural modeling~\cite{xu2022point, rakhimov2022npbg++, jena2022neural, lu2023urban}. A line of works~\cite{ost2021neural, wu2023mars, fischer2024multi} seeks to disentangle urban environment by individually modeling different scene components via graph-based representations~\cite{zhang2022graph, Velickovic2017GraphAN}, where multiple neural implicit models represent static and dynamic objects as nodes, while 3D bounding boxes and their spatial relationships define the edges. 



However, few existing works on urban scenes address the limited view distribution in urban scenes, as training images are predominantly captured from forward-facing vehicle cameras~\cite{Geiger2012CVPR, liao2022kitti}. This setup contrasts with the need for diversely posed images for accurate scene reconstruction, leading to lower quality when rendering from distant viewpoints. Existing methods~\cite{fu2022panoptic, wu2023mars, wang2024rcdn} utilize a single set of forward-facing images, restricting test viewpoints to an ``\textit{interpolative}" area. Consequently, evaluating these methods is inadequate for views that are far left, right, or downward. To the best of our knowledge, only VEGS~\cite{hwang2024vegs} optimizes scene visuals via the image diffusion paradigm while maintaining generalization for unseen views. However, the text-only conditional diffusion process is under coarse-grained control, resulting in failures of text-ambiguous or large view angles, shown in Figure~\ref{fig:intro}.

In this work, we introduce UrbanCraft, a framework designed to address the challenges of Extrapolated View Synthesis (EVS) by leveraging hierarchical semantic-geometric (sem-geometric) representations as priors. These sem-geometric priors effectively mitigate ambiguities in text prompts, providing refined control and enhancing the quality of synthesis views. Our framework excels in repairing unseen extrapolated views by leveraging the advanced generation capabilities of our pretrained 2D diffusion model, named UrbanCraft2D. Specifically, we leverage the partially observable scene to reconstruct coarse semantic and geometric primitives, establishing a coarse scene-level prior through the occupancy grid~\cite{behley2019semantickitti, tian2024occ3d, li2024sscbench} as a fundamental representation. Additionally, we integrate a fine instance-level projected rotation map to refine the details of object orientations. The UrbanCraft2D leverages ControlNet~\cite{zhang2023adding} to integrate the above conditions. Based on this, to further enhance the consistency with the observed view, we propose Hierarchical Sem-Geometric-Guided Variational Score Distillation (HSG-VSD). 
This approach allows us to control the score distillation process with 2D conditional signals derived from the occupancy grids.
These signals provide explicit priors for the semantic and geometric distribution of the scene, guiding the optimization process to align with the observable scene and ensuring high fidelity and 3D consistency in the results.

Trained with urban autonomous dataset~\cite{liao2022kitti, caesar2020nuscenes}, our work achieves state-of-the-art urban extrapolated camera view synthesis performance, both quantitatively and qualitatively. We present the \textit{first} effective framework to synthesize large- and text-ambiguous extrapolated camera views with \textit{consistent distribution} with existing camera views.

\section{Related Works}
\noindent\textbf{Neural Representations for Urban Scenes.} NeRF~\cite{mildenhall2021nerf} utilizes implicit neural representations to encode densities and colors of the scene, employing volumetric rendering for view synthesis, which can be effectively optimized from 2D multi-view images. Numerous works have enhanced NeRF regarding rendering quality~\cite{barron2021mip, barron2022mip, barron2023zip, hu2023tri}, efficiency~\cite{chen2022tensorf, fridovich2022plenoxels, kerbl20233d, muller2022instant, sun2022direct}. Notably, 3DGS~\cite{kerbl20233d}, an innovative variant of point-based rendering, advanced high-fidelity real-time rendering with point-based scene representation and its differential, rasterization-based splatting techniques. Based on 3DGS, ~\cite{Ost_2021_CVPR, yang2023unisim, zhou2024drivinggaussian, yan2024street} has explored disentangling static and dynamic components in urban scenes to enhance rendering efficiency and realism. By leveraging scene graph representations, urban relationships can be modeled explicitly, enabling better temporal consistency, motion separation, and interactive scene manipulation. Hence, in this work, we follow the static and multiple instance-wise dynamic decoupled Gaussian representations for urban scenes.



\noindent\textbf{Scene Reconstruction under Limited Viewpoints.} Recent works\cite{jain2021putting, niemeyer2022regnerf, wynn2023diffusionerf} on few-shot NeRFs define a problem where there are a few sparsely posed yet well-distributed images for training. For the extrapolated view synthesis setting, the biased distribution of train cameras is heavily emphasized rather than their number. RapNeRF~\cite{zhang2022ray} trains with densely positioned cameras at a fixed altitude and tests at different altitudes. However, it assumes view-agnostic color for unseen rays, which is unsuitable for outdoor scenes with reflective surfaces and varying lighting. NeRFVS~\cite{yang2023nerfvs} improves the approach by using holistic priors like pseudo-depth maps and view coverage from neural reconstructions, demonstrating enhanced rendering for diverse 3D indoor scenes. VEGS~\cite{hwang2024vegs} utilizes the image diffusion model to teach the visual characteristics of the scene while keeping its generalization capability for unseen views. However, the text-only conditioned diffusion is under coarse-grained control, resulting in failures of text-ambiguous or large unseen view angles. 
In this work, we tackle the above issue through the proposed UrbanCraft2D with hierarchical sem-geometric priors to achieve more fine-grained control.
\begin{figure*}[htbp]
    \centering
    \includegraphics[width=\linewidth]{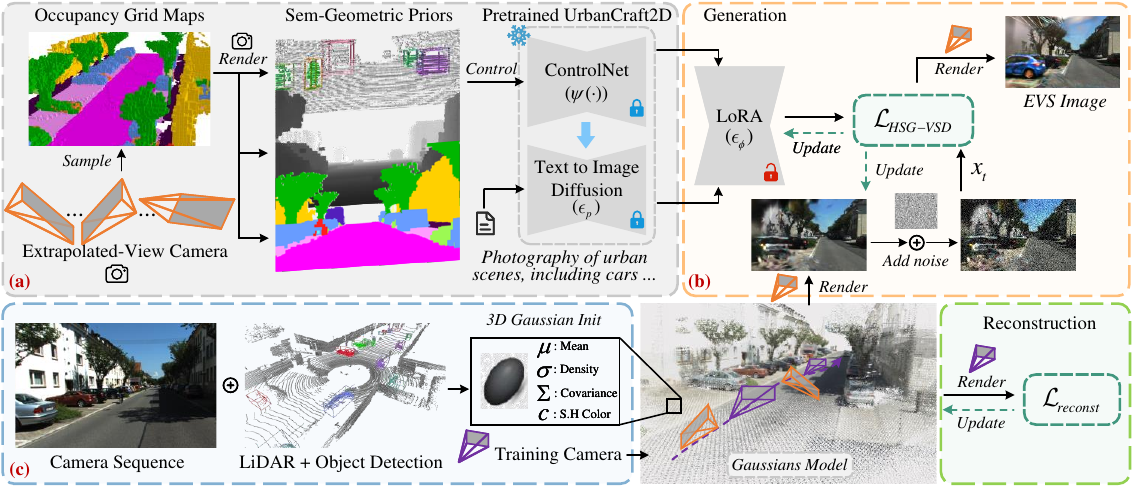}
    \caption{\textbf{Overview of UrbanCraft}. We introduce UrbanCraft, a method that repairs unseen extrapolated views with hierarchical sem-geometric priors. Our framework contains three stages: (a) pretrained of a 2D diffusion model, named UrbanCraft2D, including stable diffusion model $\epsilon_{p}$ and corresponding ControlNet $\psi(\cdot)$ and (b) distillation of the UrbanCraft2D by proposed HSG-VSD to enforce the optimization process to be consistent with the observable scene and (c) initialization for urban 3D representation.}
    \label{fig:pipeline}
\end{figure*}
\noindent\textbf{Text-to-3D.} Early text-to-3D methods employed CLIP~\cite{radford2021learning} for optimization guidance but had challenges in creating high-quality 3D content~\cite{jain2022zero, hong2022avatarclip, mohammad2022clip, sanghi2022clip, lu2024urbanArchi}. Recently, large-scale diffusion models~\cite{balaji2022ediff, rombach2022high, saharia2022photorealistic} have exhibited remarkable efficacy in text-to-image generation, adept at producing 2D images with high fidelity and diversity. ControlNet~\cite{zhang2023adding}, which integrates 2D pixel-aligned conditional signals to regulate the generation, provides more fine-grained control. Hence, The achievements in 2D content creation have propelled text-to-3D generation~\cite{chen2023fantasia3d, lin2023magic3d, metzer2023latent, poole2022dreamfusion, tang2023dreamgaussian, wang2024prolificdreamer, gao2023magicdrive}. As the key of contemporary text-to-3D methods, Score Distillation Sampling (SDS)~\cite{poole2022dreamfusion} optimizes a 3D model by synchronizing 2D images produced from various perspectives with the distribution obtained from a text-conditioned diffusion model. Nonetheless, SDS experiences challenges including excessive smoothness and over-saturation. VSD~\cite{wang2024prolificdreamer} proposes a particle-based variational framework to enhance the generation quality of SDS. Meanwhile, some methods have endeavored to accomplish scene-level generation, albeit the scales of the scenes remain considerably restricted~\cite{cohen2023set, bai2023componerf}. Current text-to-3D methods encounter difficulties in accurately representing the complex distribution of large-scale urban scenes when relying solely on text-conditioned diffusion models.
Alternative methods have investigated incrementally reconstruction and inpainting of the scenes to generate room-scale scenes~\cite{hollein2023text2room}. This paradigm, however, is vulnerable to 3D geometric consistency and presents difficulties in scaling for more extensive scenarios.

\section{UrbanCraft: Methodology}

Given a sequence of forward-facing images $\mathcal{I}^{k}$ and LiDAR point clouds $\mathcal{P}_{k}$ captured from a driving vehicle, our goal is to reconstruct an urban scene that can yield photo-realistic renderings from viewpoints beyond the training cameras' distribution. This paper refers to renderings on such views as Extrapolated View Synthesis (EVS). 

Our method, UrbanCraft, introduces a novel approach for large-scale 3D urban scene repair. The overall pipeline is illustrated in Figure~\ref{fig:pipeline}. The input to UrbanCraft consists of: i) a prompt providing a coarse description of the content in the extrapolated view scene, ii) an occupancy grid map serving as the coarse guidance for the partially observable urban scene, and iii) an extrapolated camera pose trajectory defined in the space of occupancy grid maps. 

UrbanCraft renders the occupancy grid map along the camera trajectory to construct ``hierarchical sem-geometric guidance" (HSG) as the condition for the pretrained 2D diffusion model, named \emph{UrbanCraft2D}. This model incorporates rich 2D priors by combining the capability of a large-scale text-to-image diffusion model with scene-specific information from urban scene datasets, shown in Figure~\ref{fig:pipeline}.\textcolor[rgb]{0.21,0.49,0.74}{(a)}. Built upon UrbanCraft2D, we propose Hierarchical Semantic-Geometric-Guided Variational Score Distillation (HSG-VSD) to enforce consistency between the distribution and the observable scene by integrating semantic and geometric constraints into the score distillation sampling process, shown in Figure~\ref{fig:pipeline}.\textcolor[rgb]{0.21,0.49,0.74}{(b)}. Finally, we simultaneously optimize the 3D scene representation (\eg, NeRF~\cite{mildenhall2021nerf} or 3D Gaussian Splatting~\cite{kerbl20233d}, shown in Figure~\ref{fig:pipeline}.\textcolor[rgb]{0.21,0.49,0.74}{(c)}) with known images at interpolated viewpoints and with HSG-VSD at extrapolated viewpoints.

Notably, \textit{our UrbanCraft is not affected by the characteristic of extrapolated camera views}. Unlike prior work which can only support limited extrapolated camera views, our extrapolated camera views can move freely in the reconstructed urban scenes, enabling EVS views much more outside of the camera poses in the training trajectory.

    
\subsection{UrbanCraft2D}

\begin{figure}
    \centering
    \includegraphics[width=\linewidth]{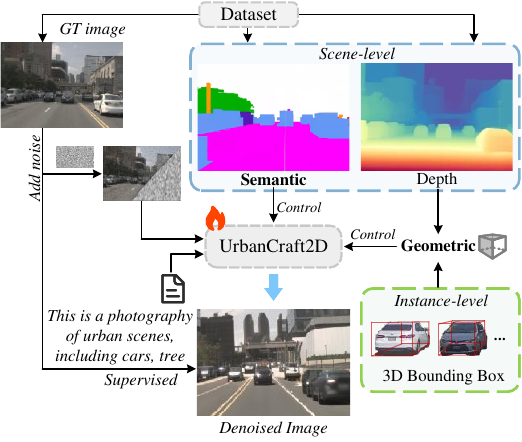}
    \caption{\textbf{UrbanCraft2D}: Diffusion pre-training process. Specifically, we utilize the voxel render~\cite{liu2020neural} to generate the scene-level control of corresponding GT images and regular the instance-level control by box2camera coordinate-based rotation maps.}
    \label{fig:urban2d-training}
\vspace{-2ex}
\end{figure}

\begin{figure}[t]
    \centering
    \includegraphics[width=\linewidth]{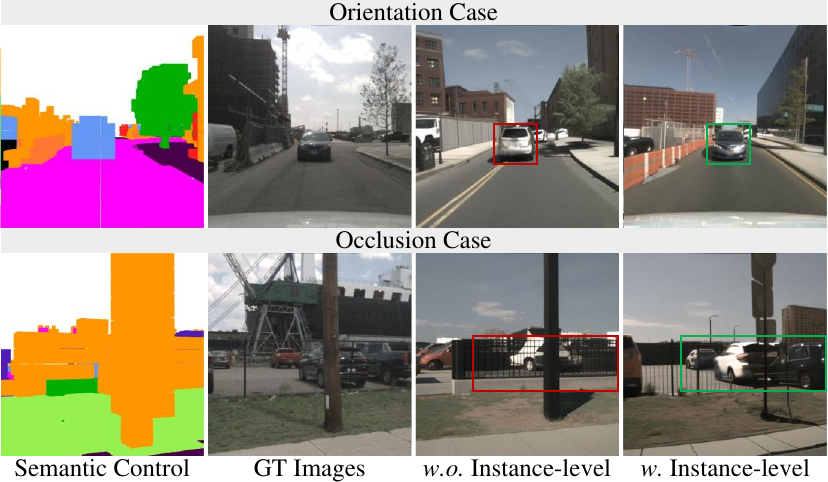}
    \caption{Illustration of the effectiveness of the proposed Hierarchical Sem-Geometric Priors for UrbanCraft2D. Note that: i) the scene-level-only control signal can guide UrbanCraft2D to generate urban scenes under reasonable distribution, and ii) adding the extra instance-level control signal enables more precise optimization of vehicles' spatial relationships.}
    \label{fig:urban2d-aba}
\vspace{-4ex}
\end{figure}

\noindent\textbf{Hierarchical Sem-Geometric Priors.} To provide an in-hand, camera view-aware control signal for large-scale urban scenes, we design the hierarchical sem-geometric representation. As shown in Figure~\ref{fig:urban2d-training}, i) scene-level priors provide coarse semantic and depth-based primitives, to make sure the general layout of urban scenes is under reasonable distribution, ii) instance-level priors provide fine geometric control via utilizing the projected object rotation map based on the extrapolated camera views, which makes the vehicles spatial relationships (\eg, numbers, occlusion or orientation) under the same 3D consistency with the observable scene, iii) as shown in Figure~\ref{fig:urban2d-aba}, the proposed hierarchical priors complementary both semantic and geometric control signals for UrbanCraft2D, achieving more accurate and high-fidelity {\color{black} controllable 2D urban scene generation}
, which is crucial for further 3D distillation.

\noindent\textbf{Pretraining.} Our UrbanCraft2D is {\color{black} a conditional version of the pretrained large-scale text-to-image diffusion model $\epsilon_{p}$}, 
with an additional HSG condition at the extrapolated camera view, which contains both the semantic and geometric primitives. {\color{black} The conditional signal consists of three parts: (1) The instance-level rotation map $\mathcal{R} \in \mathbb{R}^{H \times W \times 9}$ (assigned the pixels where the object is located to the value of a one-dimensional vector reshaped by the rotation matrix $\mathcal{M}\in\mathbb{R}^{3 \times 3}$) for instance pose control; (2) semantic map $\mathcal{S} \in \mathbb{R}^{H \times W \times 3}$ for coarse semantic control; (3) depth map $\mathcal{D} \in \mathbb{R}^{H \times W \times 1}$ for coarse geometry control, resulting in a control signal $\mathcal{C} \in \mathbb{R}^{H \times W \times 13}$. Then, $\mathcal{C}$ will be injected in the model via ControlNet~\cite{zhang2023adding} $\psi(\cdot)$, shown in Figure~\ref{fig:urban2d-training}.} Specifically, we finetune Stable Diffusion v2.1~\cite{rombach2022high} with urban scenes datasets including nuScenes~\cite{caesar2020nuscenes} and KITTI-360~\cite{liao2022kitti}. {\color{black}During the training of ControlNet, we utilized the provided occupancy grid and camera poses to generate the HSG condition.}
Meanwhile, instead of using existing caption tools such as BLIP~\cite{li2022blip} to generate prompts, we use a single base prompt for all training samples. Note that the base prompt does not need to contain any information describing the image content, it merely serves as a placeholder to avoid the model overfitting to any particular word or sentence. Specifically, we use ``\textit{This is photography of an urban street view, including cars, trees.}” as the base prompt. After finetuning, UrbanCraft2D can generate high-quality images according to the given HSG.

\subsection{Distillation-Guided Urban Scene Optimization}
To generate the extrapolated camera views, we need to distill the generation ability of our pretrained UrbanCraft2D model to urban scene reconstruction. Existing SDS or VSD-based pipelines~\cite{hwang2024vegs} rely on text prompts to guide model optimization. However, {\color{black} previous pipelines are typically used for 3D content generation in small-scale scenes, which prefers diverse generative results. However, scene reconstruction prefers deterministic results, which require the generated results to have a consistent distribution with known camera views. In such context, a single text prompt $y$ can not provide local semantic and geometric constraints of complex urban scenes, thus leading to inconsistent distribution with the existing camera views.}

Given the above observations, we propose to condition the distilling process via the proposed hierarchical sem-geometric representations rendered at the extrapolated camera pose $\mathbf{T}$. Specifically, at each generation step, we render the 2D control signals $\mathcal{C}$ from the sem-geometric representations $SG$ at the extrapolated camera view $\mathbf{T}$. Subsequently, the features produced by ControlNet $\psi(\cdot)$ are integrated into both the diffusion model $\epsilon_{p}$ and the LoRA model~\cite{hu2021lora} $\epsilon_{\phi}$, leading to a compact, scene-consistent target distribution $p_{0}(\boldsymbol{x}_{0}|y,\psi(SG(\mathbf{T})))$. Formally, the gradient of our hierarchical Semantic-Geometric-Guided Variational Score Distillation (HSG-VSD) loss can be written as:
\begin{equation}
\resizebox{0.90\hsize}{!}{$
\begin{split}
    \label{eq:self}
    \nabla_{\theta}\mathcal{L}_{\mathrm{HSG-VSD}}(\theta) & \triangleq \mathbb{E}_{t, \epsilon, \mathbf{T}}[\omega(t)(\epsilon_{p}(\boldsymbol{x}_{t},t,y,\psi(SG(\mathbf{T}))-\\
    &\epsilon_{\phi}(\boldsymbol{x}_{t}, t, \mathbf{T}, y, \psi(SG(\mathbf{T}))))\frac{\partial \boldsymbol{g}(\theta, \mathbf{T})}{\partial \theta}],
\end{split}$}
\end{equation}
where random noise $\boldsymbol{\epsilon} \sim \mathcal{N}(\boldsymbol{0}, \boldsymbol{\textit{I}})$, $t \sim \mathcal{U}(0.02,0.98)$, $\omega(t)$ weights the loss given the time step $t$, and $\boldsymbol{x}_{t}$ is the randomly perturbed rendered image.

Except for the HSG-VSD loss, we also use an additional SDS loss (\textit{i.e}, $\mathcal{L}_{\mathrm{G-SDS}}$) to refine the extrapolated scene geometry. Concretely, we encourage the rendered normal and depth images to be score distillation sampled by the general diffusion model, resulting in a better geometry distribution of urban scenes. Finally, the two losses jointly optimize the extrapolated scene, forcing the 3D representation $\theta$ to align with the semantic and geometric distribution of the observable scenes.


\subsection{Urban 3D Representation Initialization}
We leverage 3D Gaussian Splatting (3DGS)~\cite{kerbl20233d} to construct a dynamic 3D scene representation, which includes both static and multiple instance-wise dynamic decoupled Gaussian models. 

Furthermore, we have learned~\cite{qi2018geonet, yan2023nerf, han2024extrapolatedurbanviewsynthesis} that optimizing Gaussian models with forward-facing cameras causes the covariance shapes of the Gaussians to overfit to a particular perspective. We postulate that this overfitting occurs because the covariance is trained to encompass the frustum of a training pixel with minimal optimization effort. Consequently, these covariances are prone to generate undesirable cavities on the underlying scene surface,become apparent when viewed from unobserved angles. This makes the scene representation unsuitable for the extrapolated view synthesis problem.

Following \cite{hwang2024vegs}, we utilize the surface normal priors to guide covariance orientation and shape. Specifically, we employ the covariance rendering technique to approximate the scene surface normal from the rendered covariance map. Then, we supervise the map with a surface normal estimated~\cite{eftekhar2021omnidata, kar20223d} from training images, with squared L2 distance loss $\mathcal{L}_{\mathrm{normal}}$.

\subsection{Optimization}

For the optimization of UrbanCraft, the total loss is the weighted combination of the reconstruction loss and distillation-guided loss, which can be formalized as follows:
\begin{gather}
    \mathcal{L}_{\mathrm{reconst.}} = (1 - \lambda_{r})\mathcal{L}_{1} + \lambda_{r}\mathcal{L}_{\mathrm{D-SSIM}} + \mathcal{L}_{\mathrm{normal}} \\
    \mathcal{L}_{\mathrm{Distill.}} = \mathcal{L}_{\mathrm{HSG-VSD}} + \mathcal{L}_{\mathrm{G-SDS}} \\
    \mathcal{L} = \lambda_{1}\mathcal{L}_{\mathrm{reconst.}} + \lambda_{2}\mathcal{L}_{\mathrm{Distill.}},
\end{gather}
where the $\lambda_{1}, \lambda_{2}, \lambda_{r}$ is set to $1e4$, $1.0$, $0.2$, respectively.

\begin{figure*}[htbp]
    \centering
    \includegraphics[width=\linewidth]{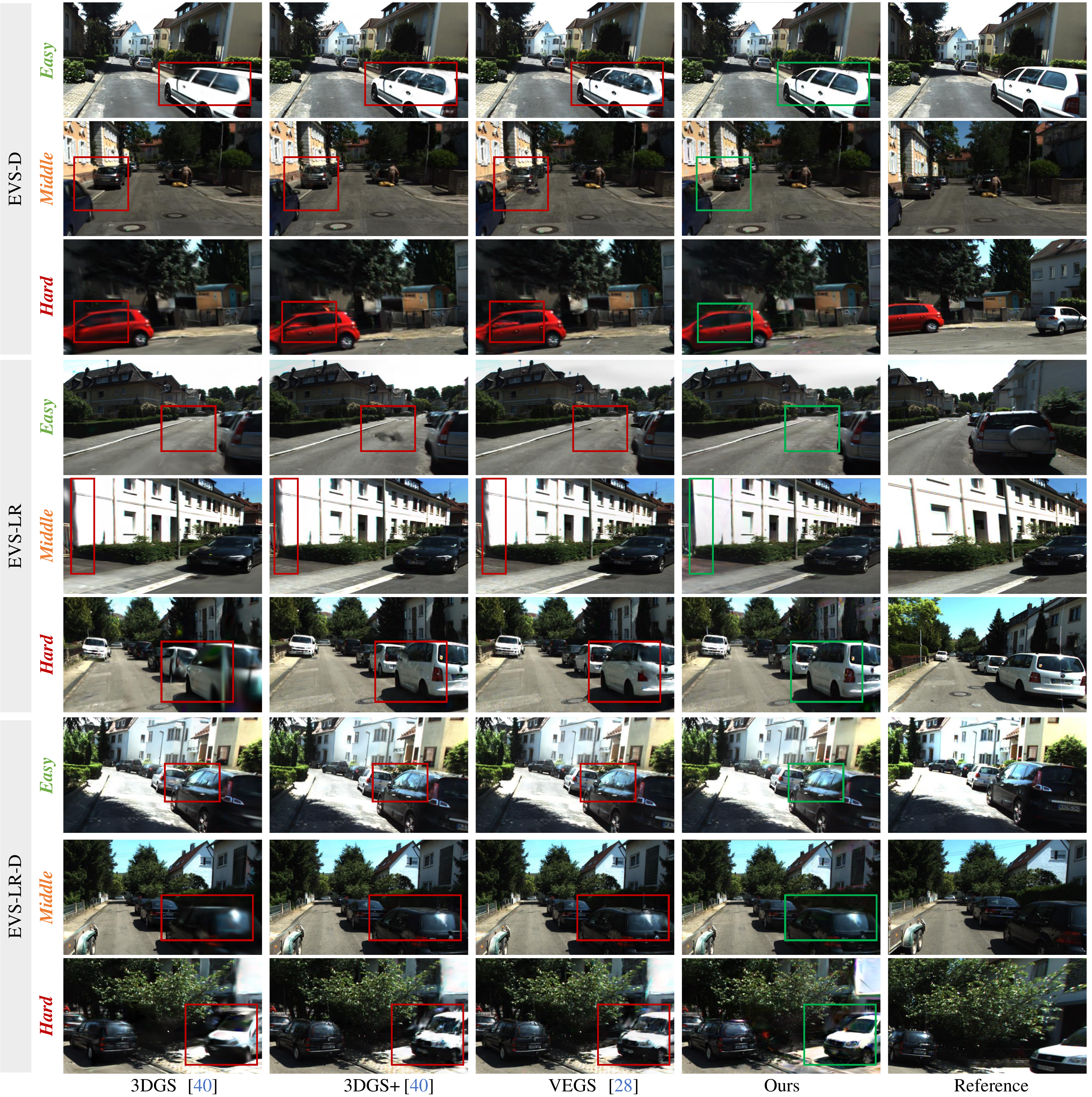}
    \vspace{-6mm}
    \caption{Qualitative comparison on KITTI-360~\cite{liao2022kitti} for extrapolated view synthesis under different difficulty levels ({\color[RGB]{46,139,87}{Easy}}, {\color[RGB]{208,122,64}{Middle}} and {\color[RGB]{255,0,0}{Hard}}) across three settings. EVS-D and EVS-LR refer to extrapolated views facing downwards and left/right, respectively, while EVS-LR-D represents a combination of both. Our method effectively reconstructs foreground and background regions, preserving object structures and scene consistency.  We also report training images for reference that maximally cover the view space of EVS from another location for comparison. Notably, our proposed UrbanCraft outperforms the baselines regarding geometry and visual sanity. 
    }
    \label{fig:results-total}
    \vspace{-2ex}
\end{figure*}

\begin{figure*}[!t]
    \centering
    \includegraphics[width=\linewidth]{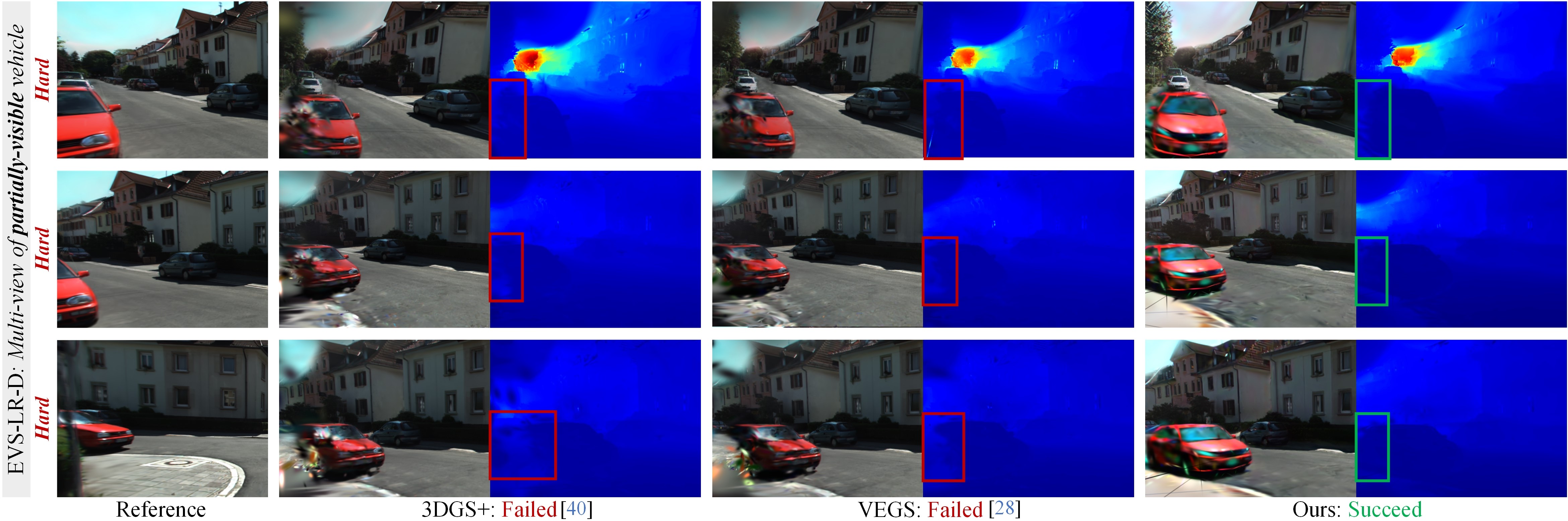}
    \vspace{-4ex}
    \caption{Qualitative comparison about extrapolated view synthesis of multi-view distribution consistency of partially visible vehicle. Ours effectively synthesizes large extrapolated views while maintaining both semantic and geometric distribution consistency with the observed scene. (i) In terms of \textbf{semantic consistency}: our method preserves the red color consistency of the vehicle, and the synthesized front part exhibits semantic symmetry with the original vehicle. (ii) In terms of \textbf{geometry consistency}: our method accurately completes the missing depth of the vehicle while ensuring smooth continuity between the synthesized and observed vehicle.
    }
    \label{fig:results}
    \vspace{-2ex}
\end{figure*}

\section{Experiments}
\subsection{Experimental Setup}
\label{sec41}
\noindent\textbf{Dataset and Implementation Details.} We conduct our experiments on KITTI-360~\cite{liao2022kitti} and NuScenes~\cite{caesar2020nuscenes} Datasets, which offer 3D bounding box annotations, forming extensive 3D scene information. Then, we utilize the SSCBench~\cite{li2024sscbench} and Occ3D~\cite{tian2024occ3d} to prepare the corresponding occupancy grid maps. We utilize  Stable Diffusion v2.1~\cite{rombach2022high} as the text-to-image diffusion model. The overall framework is implemented using PyTorch~\cite{paszke2019pytorch}, and we utilize the diffuser~\cite{von-platen-etal-2022-diffusers} to implement the proposed UrbanCraft2D. For the training of our UrbanCraft2D, we finetune with our produced hierarchical sem-geometric prior data. Specifically, we concatenate the rendered semantic, depth and rotation maps from occupancy grid maps as the input conditional signal. To adapt the Stable Diffusion v2.1, we crop and resize the signal into a $512 \times 512$ resolution. The AdamW~\cite{Loshchilov2017DecoupledWD} is utilized as the optimizer with an init learning rate of $1 \times 10^{-3}$. The training of a single sequence is conducted on a single NVIDIA A800 GPU.

\noindent\textbf{Extrapolated Cameras.} To ensure fairness, our experimental design follows previous work~\cite{hwang2024vegs} and extent it to a more comprehensive setting. We construct three EVS camera sets: EVS-D, EVS-LR and EVS-LR-D, each featuring three difficulty levels ({\color[RGB]{46,139,87}Easy}, {\color[RGB]{208,122,64}Middle}, {\color[RGB]{255,0,0}Hard}). EVS-D is created by rotating the test cameras $10^{\circ}$ around the X-axis of the camera coordinate (pointing to the right) and translating them upward by $\color[RGB]{46,139,87}{0.2}$/$\color[RGB]{208,122,64}{0.4}$/$\color[RGB]{255,0,0}{0.6}$ meters in world units. EVS-LR involves rotating the test cameras $\color[RGB]{46,139,87}{\le\!15^{\circ}}$, $\color[RGB]{208,122,64}{15^{\circ}\!-\!45^{\circ}}$ and $\color[RGB]{255,0,0}{\ge\!45^{\circ}}$ around the Z-axis of the upward-pointing world coordinate to look left and right. EVS-LR-D integrates both EVS-LR and EVS-D variations at each difficult level, respectively. {\color{red} Please refer to Supplementary Material for discussion of performance under highly challenging scenarios.}



\noindent\textbf{Baselines.} We made our own baseline using VEGS~\cite{hwang2024vegs}, an existing state-of-the-art method for urban EVS problem. We also compare with 3DGS~\cite{kerbl20233d} to compare the relative performance between the state-of-the-art point-based rendering methods. Meanwhile, to make 3DGS more suitable for dynamic urban scenes, we include LiDAR initialization and static-dynamic decoupled modeling as 3DGS+. For quantitative evaluation, to ensure fairness, we utilize the same evaluation metrics Fr\'echet Inception Distance (FID)~\cite{heusel2017gans} and Kernel Inception Distance (KID)~\cite{binkowski2018demystifying}, which are computed with 4k randomly sampled images.

\subsection{Qualitative Evaluation}
\noindent\textbf{Comparison to Baselines.} In Figure~\ref{fig:results-total}, we demonstrate qualitative results of UrbanCraft on the KITTI-360 dataset, highlighting its superior performance compared to baseline methods in EVS-D, EVS-LR and EVS-LR-D settings at three difficult levels. These results clearly illustrate the strengths of our approach in handling complex extrapolated view synthesis tasks. Specifically, we observe the following: i) \textbf{Overall performance}: UrbanCraft effectively reconstructs both foreground and background regions, maintaining object structures and scene consistency. Meanwhile, UrbanCraft significantly outperforms previous methods, with a notable advantage over VEGS. The latter struggles to accurately model large and ambiguous extrapolated views due to the coarse-grained control inherent in its text-only diffusion model, and ii) \textbf{EVS-D setting}: UrbanCraft enhances both semantic and geometric details by leveraging the pretrained UrbanCraft2D model, utilizing the proposed HSD-VSD distillation to refine the quality of view synthesis quality, and iii) \textbf{EVS-LR setting}: Extrapolated views in this setting often exhibit under-reconstructed regions at the edges of the frame, a common issue when capturing urban scenes with forward-facing cameras. Baseline methods suffer from artifacts such as black holes (EVS-LR \textit{{\color[RGB]{46,139,87}Easy} level}) and blurring (EVS-LR \textit{{\color[RGB]{255,0,0}Hard} level}). In contrast, UrbanCraft effectively synthesizes these areas, demonstrating resilience to the characteristics of extrapolated camera views by incorporating hierarchical semantic-geometric priors, and iv) \textbf{EVS-LR-D setting}: Especially for \textit{{\color[RGB]{255,0,0}Hard} level}, the baseline methods all suffer from a collapse phenomenon in rendering the white car, leading to an unstable geometric structures. In contrast, UrbanCraft successfully preserves the car's geometric integrity, ensuring a more coherent reconstruction.

\begin{table}[!t]
\centering
\caption{Quantitative comparison on the KITTI-360. FID~\cite{heusel2017gans} and KID~\cite{binkowski2018demystifying} are measured between EVS and training images. PSNR, SSIM and LPIPS are measured from conventional test cameras on static scenes where ground-truth images are available.}
\label{table1}
\vspace{-1mm}
\setlength\tabcolsep{7.5pt}
\footnotesize
\begin{tabular}{c|cc|ccc}
\hline
Methods & FID$\downarrow$ & KID$\downarrow$   & PSNR$\uparrow$ & SSIM$\uparrow$ & LPIPS$\downarrow$ \\ \hline
3DGS    & 321.61 & 1.0038 &  23.55    & 0.809     & 0.265      \\
3DGS+   & \cellcolor[RGB]{211,210,210}279.82 & \cellcolor[RGB]{211,210,210}0.9137 &  \cellcolor[RGB]{215,240,208}24.16    &  \cellcolor[RGB]{215,240,208}0.834    &  \cellcolor[RGB]{215,240,208}0.207     \\
VEGS    & \cellcolor[RGB]{214,230,242}279.27 & \cellcolor[RGB]{214,230,242}0.9245 &  \cellcolor[RGB]{211,210,210}24.00    &   \cellcolor[RGB]{214,230,242}0.830   &   \cellcolor[RGB]{214,230,242}0.216    \\ \hline
Ours    & \cellcolor[RGB]{215,240,208}278.75 & \cellcolor[RGB]{215,240,208}0.8906 &  \cellcolor[RGB]{214,230,242}24.15    &   \cellcolor[RGB]{214,230,242}0.830   &   \cellcolor[RGB]{211,210,210}0.219    \\ \hline
\end{tabular}
\vspace{-3mm} 
\end{table}

\begin{figure}[t]
    \centering
    \includegraphics[width=\linewidth]{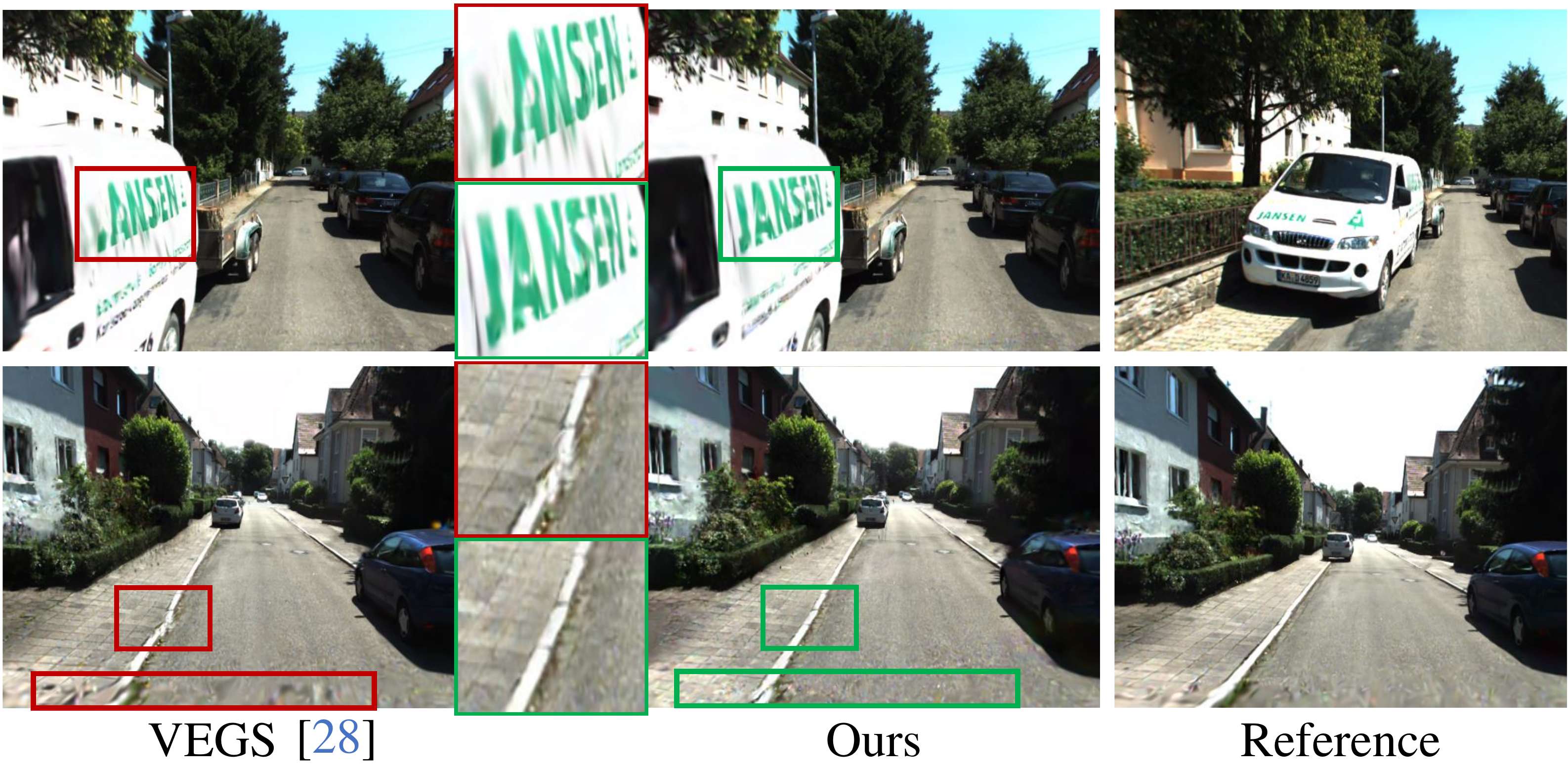}
    \vspace{-6mm}
    \caption{Fine-grained comparison with VEGS. Note that ours can refine the semantic and geometric distribution of EVS more than VEGS via more fine-grained control priors.}
    \label{fig:results_details}
    \vspace{-4mm}
\end{figure}

\noindent\textbf{Fine-grained Comparison with VEGS.}  In Figure~\ref{fig:results_details}, we showcase detailed renderings results from both VEGS and our UrbanCraft. 
Although VEGS also leverages the scene knowledge priors of a large-scale diffusion model, it relies solely on coarse text as the conditional input and lacks descriptions of the scene’s semantic and geometric details. 
As a result, in scenes with text ambiguity or significantly extrapolated camera views, as shown in Figure~\ref{fig:results_details}, the scene priors in VEGS may lose effectiveness.
In contrast, UrbanCraft leverages the proposed hierarchical sem-geometric priors as additional control signals to supplement the text prompts, ensuring that the scene priors guide novel synthesis consistently across all extrapolated camera views.


\noindent\textbf{Challenge Task: Multi-view of Partially Visible Vehicle.} To assess the multi-view distribution consistency of different methods, we present RGB and depths renderings from both the baseline methods and our proposed UrbanCraft under the same continuous extrapolated camera trajectory in Figure~\ref{fig:results}. 
Due to the lack of sufficient views to achieve geometric consistency in extrapolated views, all baseline methods fail to accurately reconstruct vehicles along the continuous extrapolated camera trajectories. As shown in Figure~\ref{fig:results}, we can observe that: i) for the semantic aspect, UrbanCraft maintains color consistency with the observable vehicle (red in Figure~\ref{fig:results}), while ensuring that the synthesized vehicle’s headstock exhibits semantic symmetry with the original vehicle, and ii) for the geometric aspect, UrbanCraft successfully synthesizes the unseen extrapolated views of red vehicles with high geometric consistency and completes the missing depth of the vehicle.

\subsection{Quantitative evaluation}

We present quantitative comparisons with baseline methods in Table~\ref{table1}. We follow VEGS~\cite{hwang2024vegs} to employ FID and KID scores with respect to the training images to assess the reconstruction quality of EVS renderings. Due to the significant differences in camera distribution between training images and EVS renders, as shown in Figure~\ref{fig:results}, it is unrealistic to expect small FID/KID scores. Therefore,  we interpret these scores as approximate indicators of visual coherence and similarity to the original scene. We also measure PSNR, SSIM and LPIPS~\cite{zhang2018unreasonable} to evaluate renderings on the conventional test cameras. Ours outperforms 3DGS across all metrics. While ours outperform VEGS in PSNR, VEGS performs slightly better in LPIPS, indicating on-par performance on conventional test cameras. However, ours surpasses VEGS in FID and KID measured from three EVS settings, which aligns with the analysis from the qualitative results. Additionally, comparison with 3DGS+ highlights \textit{that reconstruction quality on the conventional test cameras does not necessarily correspond to the quality on EVS.}

\subsection{Ablation Study}

We validate two key components of our framework: the instance-level control in UrbanCraft2D and the additional SDS-based $\mathcal{L}_{\mathrm{G-SDS}}$ loss. More ablation results are provided in our \emph{Supplementary Material}.

\begin{figure}[!t]
    \centering
    \includegraphics[width=\linewidth]{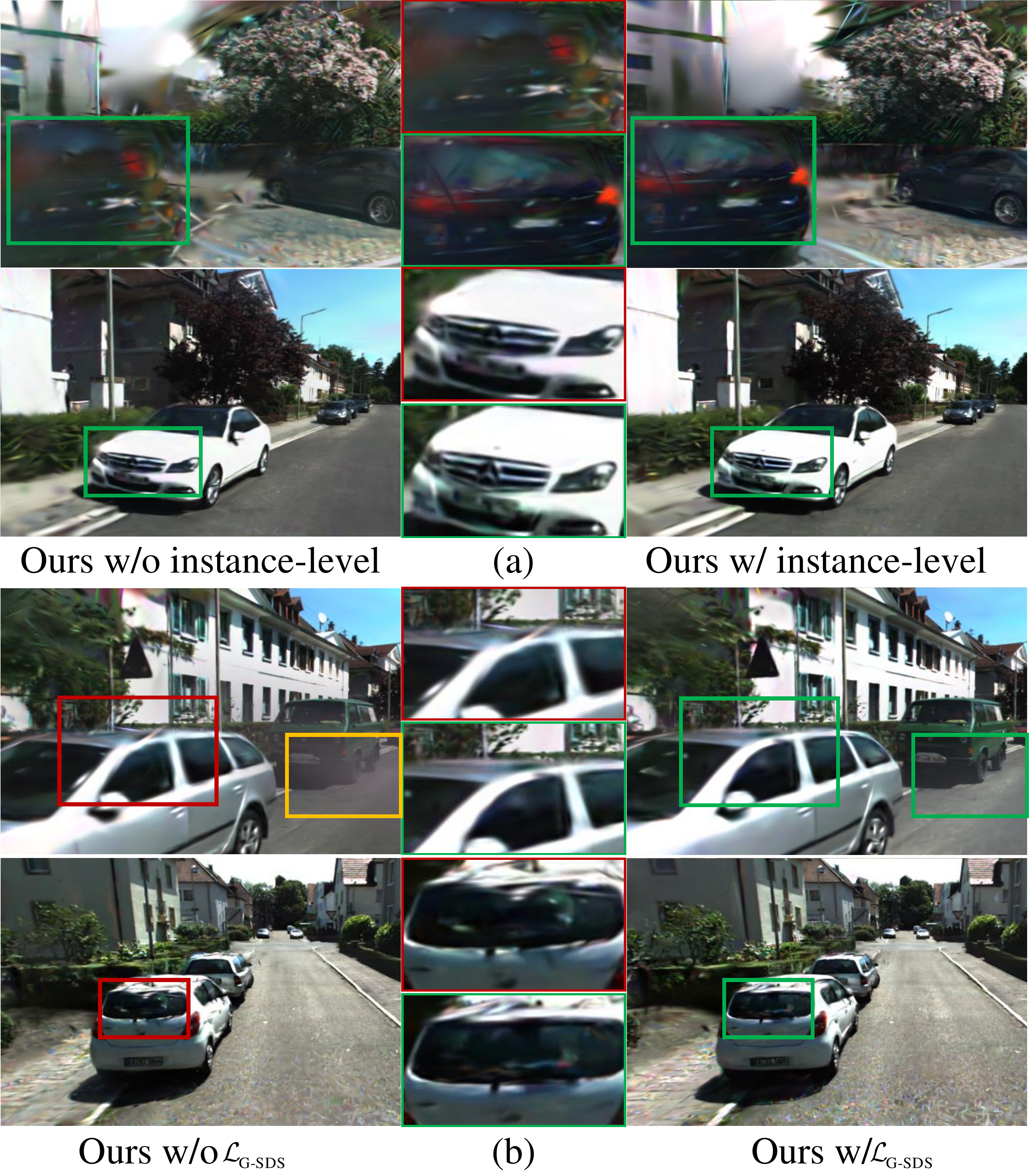}
    \vspace{-6mm}
    \caption{Qualitative ablation comparisons results on the proposed UrbanCraft2D's instance-level control and $\mathcal{L}_{\mathrm{G-SDS}}$. The instance-level control effectively guides UrbanCraft2D to learn accurate vehicles' spatial relationships, generating the broken geometry and refining the textures. $\mathcal{L}_{\mathrm{G-SDS}}$ effectively removes floating artifacts and ameliorates the sharp cone phonemes during the Gaussian-based reconstruction process.}
    \label{fig:results_ablat}
    \vspace{-4mm}
\end{figure}

\noindent\textbf{Effect of Instance-level Control.} We investigate the effectiveness of fine-grained geometric control within the proposed hierarchical sem-geometric representations. In Figure~\ref{fig:results_ablat}.\textcolor[rgb]{0.21,0.49,0.74}{(a)}, we observe that i) the accurate spatial relationships encoded in the instance-level prior from the pretrained UrbanCraft2D allow HSG-VSD to correct both the vehicles' broken semantics and their corresponding geometry, and ii) for observable scenes, with the instance-level prior, UrbanCraft effectively leverages the large-scale text-to-image diffusion model’s capability to refine the vehicles' textures and alleviate blur.

\noindent\textbf{Effect of SDS-based Geometry Loss.} As shown in Figure~\ref{fig:results_ablat}.\textcolor[rgb]{0.21,0.49,0.74}{(b)}, we observe that i) in the red bounding box section, the proposed $\mathcal{L}_{\mathrm{G-SDS}}$ effectively reduces the cavities on the vehicles' surface and ameliorate the sharp cone phenomenon, resulting in much smoother vehicle geometry, and ii) in the yellow bounding box section, the original Gaussian optimization process may produce floating artifacts due to insufficient regularization constraints. To address this, the proposed $\mathcal{L}_{\mathrm{G-SDS}}$ leverages the adequate priors from the large-scale text-to-image diffusion model to better regularize the corresponding geometry distribution. For a detailed geometric comparison, please kindly refer to our \emph{Supplementary Material}.





\section{Conclusion}
We introduce UrbanCraft, an innovative method for improving the Extrapolated View Synthesis (EVS) given training images from forward-facing cameras. The core of our method lies in incorporating the proposed hierarchical sem-geometric (HSG) representations as robust and fine-grained control signals for the pretrained 2D diffusion mode, named UrbanCraft2D. The UrbanCraft2D processes rich 2D prior by combining the capabilities of the large-scale text-to-image diffusion model and scene-specific information from the HSG representations. Built upon UrbanCraft2D, we propose the Hierarchical Sem-geometric-Guided Variational Score Distillation (HSG-VSD), which integrates the semantic and geometric constraints into the score distillation sampling process. This flexible representation, combined with the inherent capability of text-to-image diffusion models, enables our pipeline to support instance-level style editing. Experimental results demonstrate the superiority of our proposed UrbanCraft over current state-of-the-art methods, highlighting its ability to repair extrapolated views and maintain distribution consistency across the observable scenes.

{\small
\bibliographystyle{unsrt}
\bibliography{egbib}
}

\maketitlesupplementary
\tableofcontents

\section{Implementation Details.}

\subsection{UrbanCraft2D Pretraining Process} We pretrained the proposed UrbanCraft2D using both the NuScenes~\cite{caesar2020nuscenes} and KITTI-360~\cite{liao2022kitti} datasets. Experimental results demonstrate that UrbanCraft2D trained on NuScenes performs better than when trained on KITTI-360. This is because KITTI-360 provides only forward-facing images, whereas NuScenes offers images from six different viewing angles, covering a much broader distribution. Consequently, for large extrapolated camera views, prior knowledge from UrbanCraft2D pretrained on NuScenes can be better utilized and further refined through the proposed Hierarchical Semantic-Geometric-Guided Variational Score Distillation (HSG-VSD) process.

The training was conducted on a single NVIDIA A800-80G GPU with a batch size of 32 for 20 epochs, taking approximately five days to complete. Notably, when UrbanCraft2D pretrained on NuScenes is applied for inference on KITTI-360 (with UrbanCraft2D’s parameters frozen), we can address the style differences between the two datasets by fine-tuning a LoRA~\cite{hu2021lora} to achieve style transfer. 

\begin{figure}[t]
    \centering
    \includegraphics[width=\linewidth]{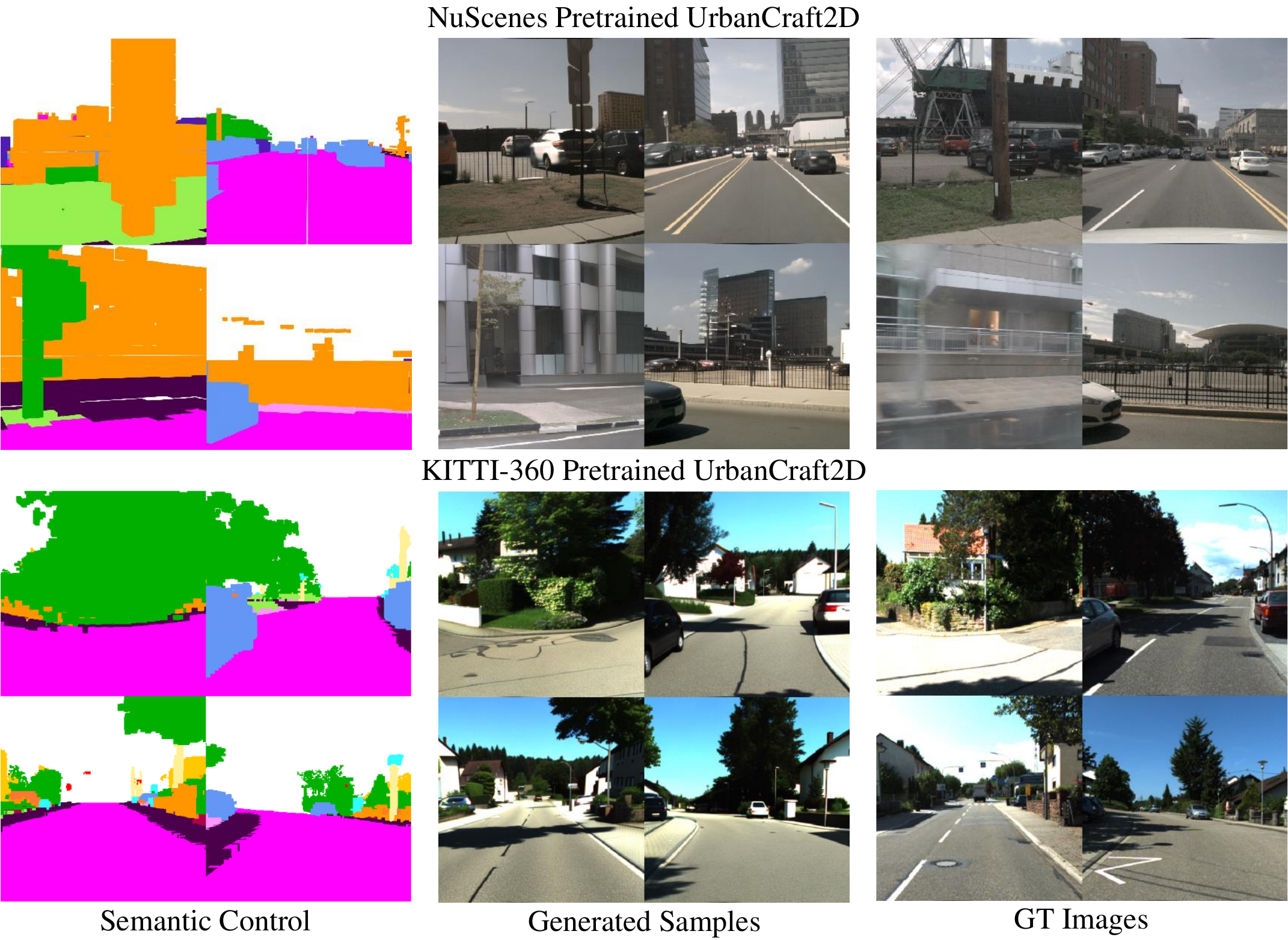}
    \vspace{-4mm}
    \caption{Qualitative comparison about NuScenes~\cite{caesar2020nuscenes} and KITTI-360~\cite{liao2022kitti} pretrained UrbanCraft2D, respectively.}
    \label{fig:supp1}
    \vspace{-4mm}
\end{figure}

\begin{figure}[t]
    \centering
    \includegraphics[width=\linewidth]{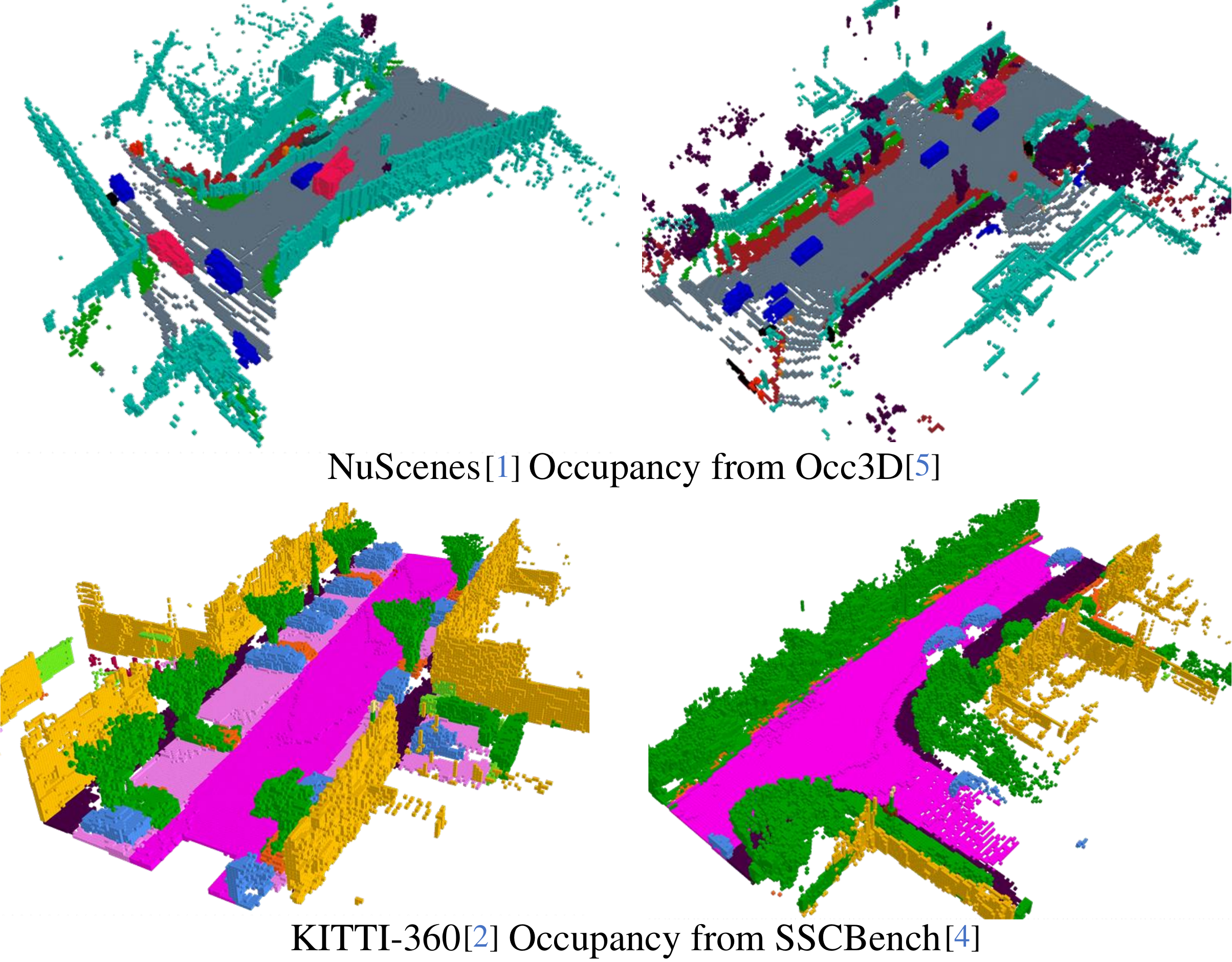}
    \vspace{-4mm}
    \caption{Visualization of occupancy grid maps.}
    \label{fig:supp4}
    \vspace{-4mm}
\end{figure}

\begin{figure*}[t]
    \centering
    \includegraphics[width=\linewidth]{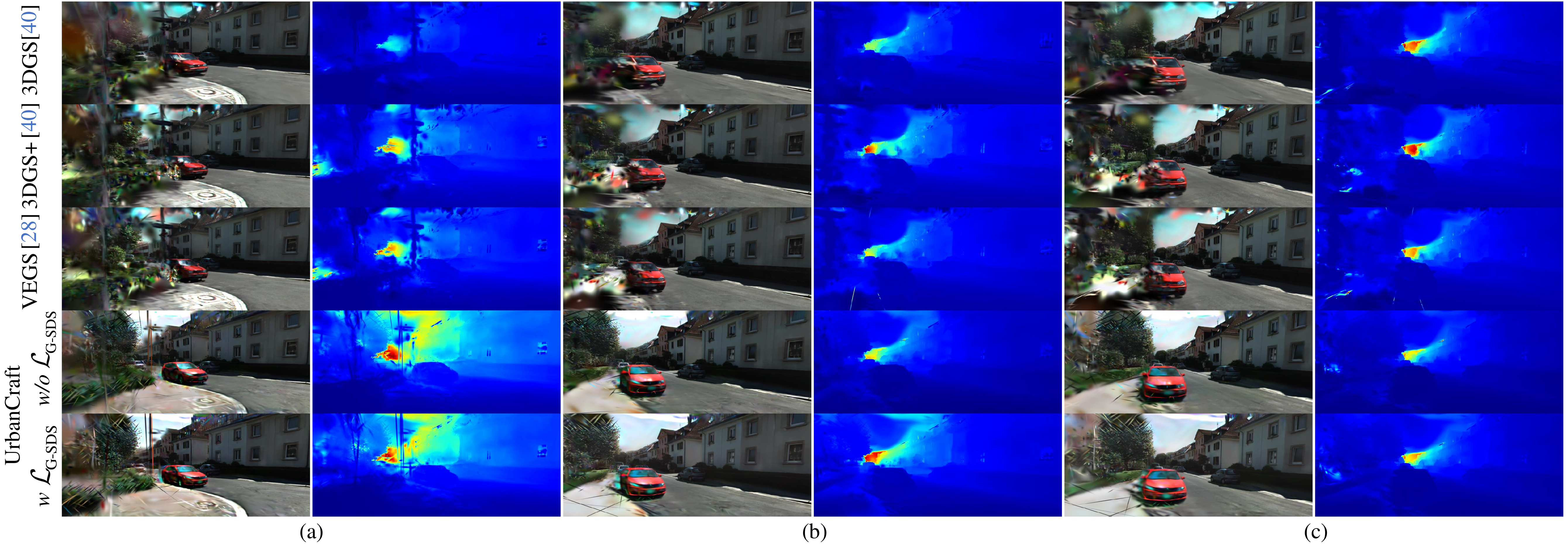}
    \vspace{-6mm}
    \caption{Qualitative comparison about multi-view distribution consistency for extrapolated view synthesis.}
    \label{fig:supp3}
    \vspace{-4mm}
\end{figure*}

\begin{figure}[t]
    \centering
    \includegraphics[width=\linewidth]{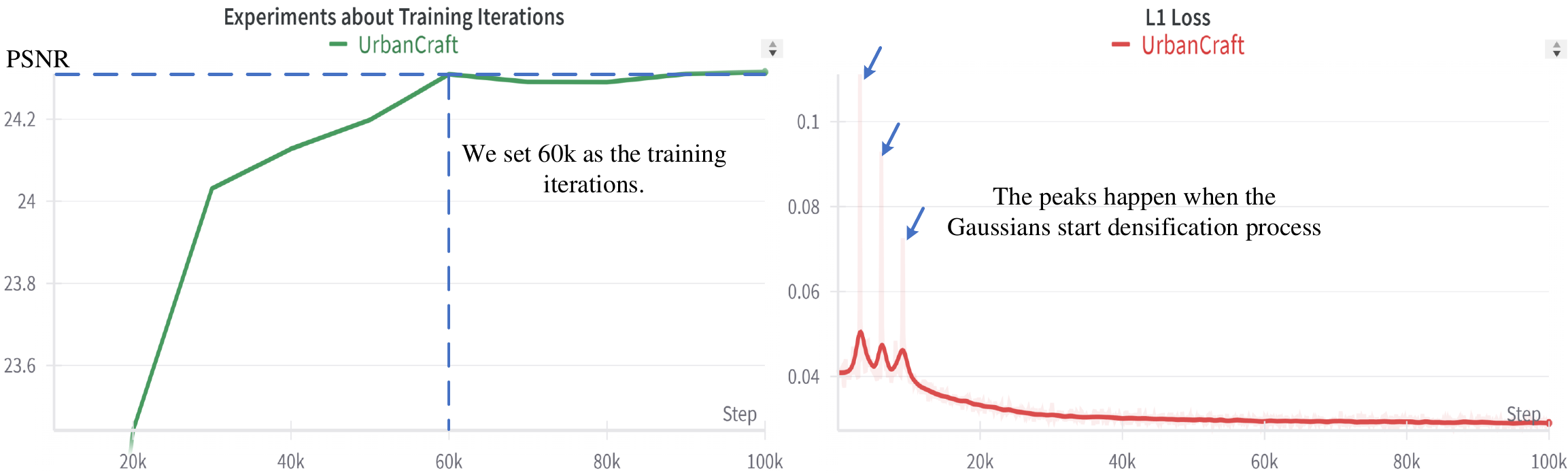}
    \vspace{-4mm}
    \caption{Hyper-parameters setting about the training iterations.}
    \label{fig:supp2}
    \vspace{-2mm}
\end{figure}


\begin{figure*}[t]
    \centering
    \includegraphics[width=\linewidth]{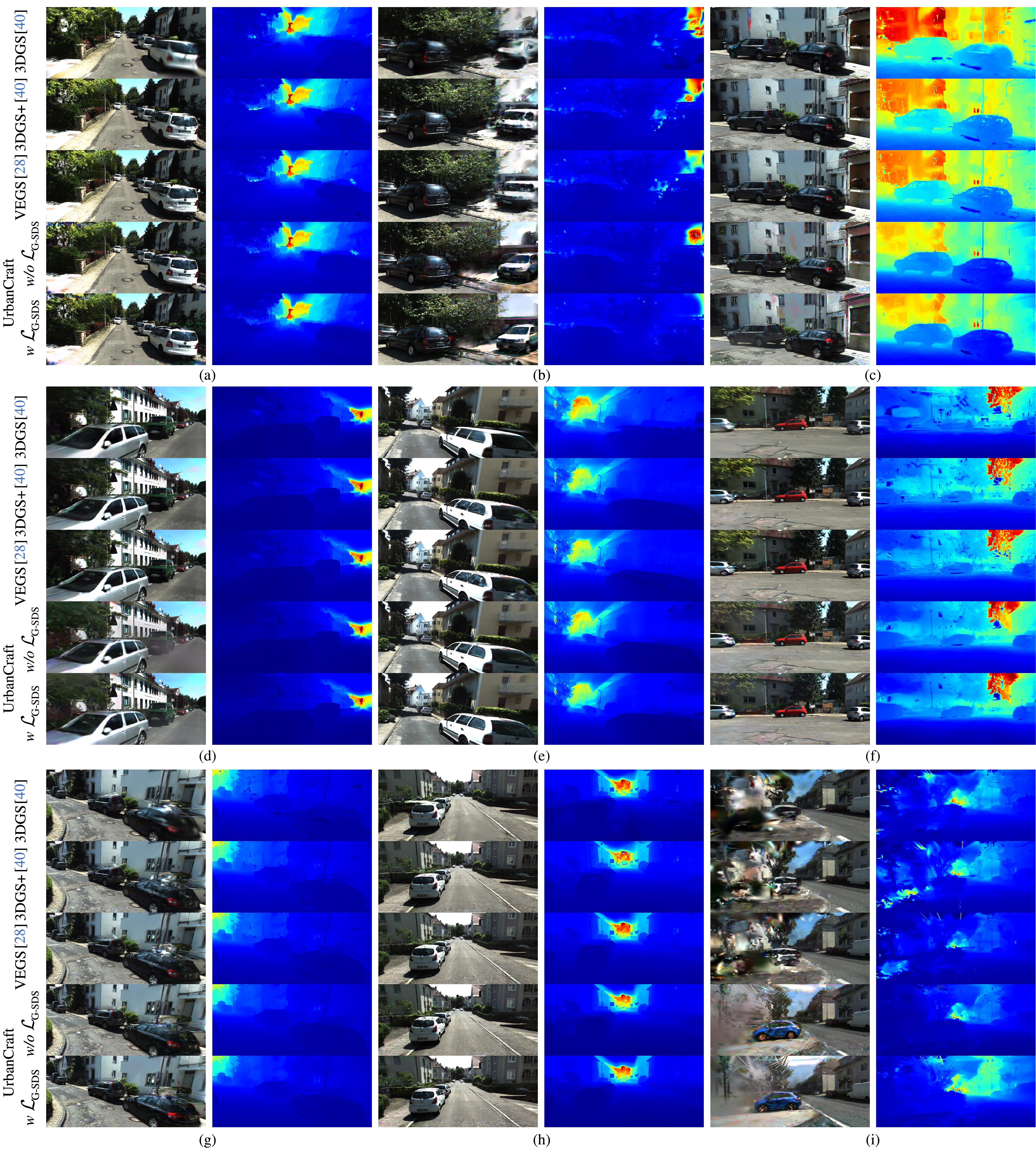}
    \vspace{-6mm}
    \caption{Qualitative comparison of RGB images and corresponding depth maps on KITTI-360~\cite{liao2022kitti} for extrapolated view synthesis.}
    \label{fig:supp5}
    \vspace{-4mm}
\end{figure*}

\begin{figure*}[t]
    \centering
    \includegraphics[width=\linewidth]{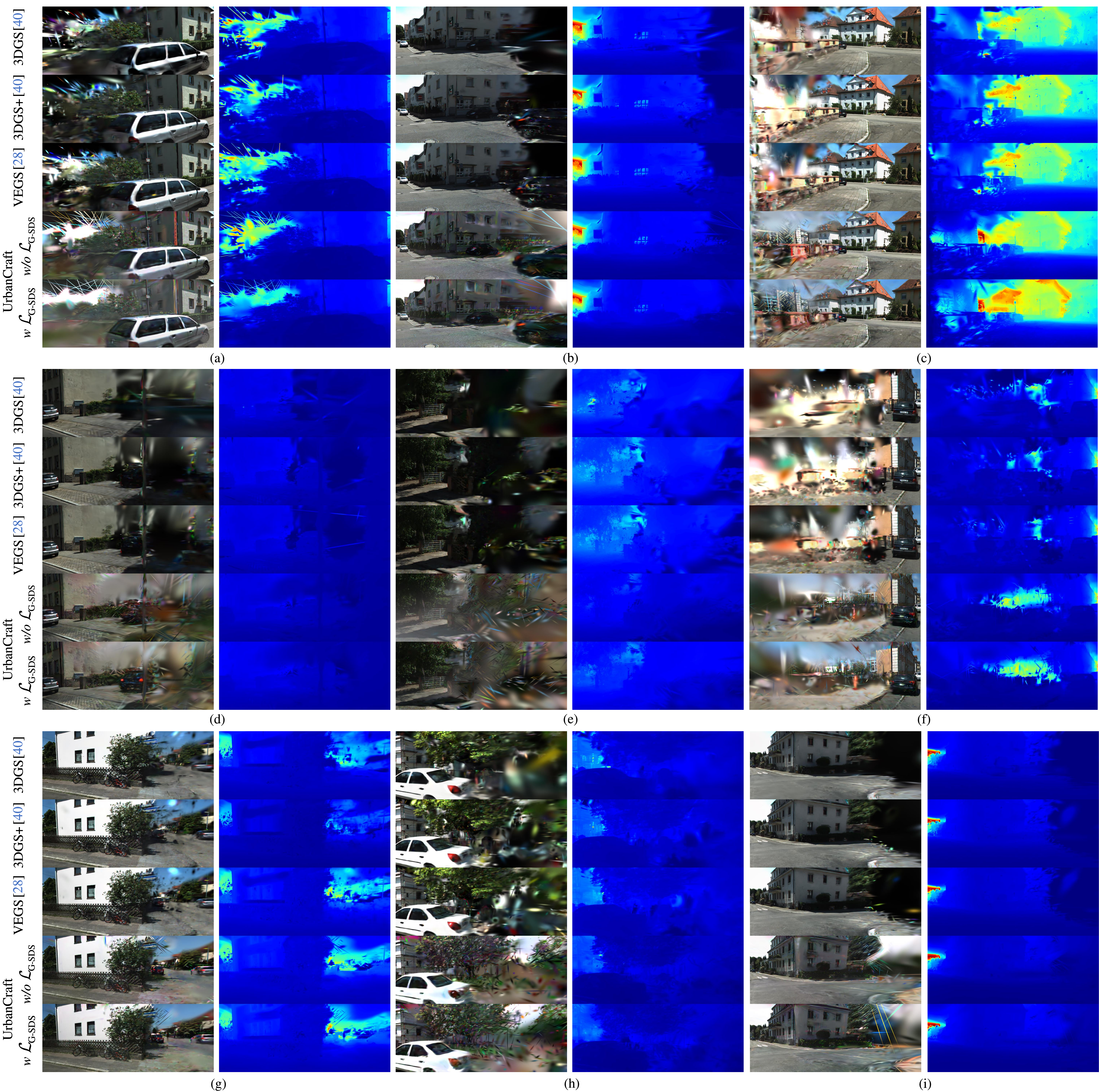}
    \vspace{-6mm}
    \caption{Qualitative comparison about failure cases for extrapolated view synthesis.}
    \label{fig:supp7}
    \vspace{-4mm}
\end{figure*}

\subsection{Occupancy Grid Maps} 
For occupancy grids, we leverage SSCBench~\cite{li2024sscbench} and Occ3D~\cite{tian2024occ3d} to provide coarse scene-level information. SSCBench is used for the KITTI-360 dataset, with a voxel size of 0.2 meters and a perception range of (51.2m, 51.2m, 6.4m). Occ3D is employed for the NuScenes dataset, with the same voxel size of 0.2 meters and a perception range of (80m, 80m, 6.4m). Notably, current 3D semantic scene completion algorithms~\cite{cao2022monoscene,li2023voxformer, jiang2023symphonies}, based on multi-sensor inputs, can generate highly accurate occupancy grid information. In principle, users can apply these completion algorithms to generate occupancy grids for custom datasets and then utilize the proposed UrbanCraft framework to enhance and repair unseen extrapolated camera views. This approach significantly improves the generalizability of urban reconstruction applications.

\subsection{Hyper-parameters Setting} 

The training process of UrbanCraft consists of two stages: i) \textbf{Initial Reconstruction}: In this stage, we focus on the coarse reconstruction of urban 3D representations $\theta$ using only the reconstruction loss $\mathcal{L}_{\mathrm{reconst.}}$. The primary goal is to provide accurate geometric and semantic initialization for the subsequent generative process. The specific Gaussian optimization parameters for this stage are as follows.
\begin{itemize}
    \item densification\_interval = 100
    \item opacify\_reset\_interval = 3\_000
    \item densify\_from\_iter = 500
    \item densify\_until\_iter = 15\_000
    \item densify\_grad\_threshold = 0.0002
\end{itemize}
ii) \textbf{Fine Reconstruction and Repair}: In the second stage, we sample extrapolated camera views and refine the scene using a combination of reconstruction loss $\mathcal{L}_{\mathrm{reconst.}}$ and distillation loss $\mathcal{L}_{\mathrm{Distill.}}$. This step further repairs the scene to improve its consistency and realism. The Gaussian optimization parameters for this stage are detailed below.
\begin{itemize}
    \item densification\_interval = 2\_000
    \item opacify\_reset\_interval = 3\_000
    \item densify\_from\_iter = 1\_000
    \item densify\_until\_iter = 10\_000
    \item densify\_grad\_threshold = 0.0002
\end{itemize}
It is worth noting: i) We also experimented with combining the two stages into a single unified process. However, this approach failed because the gradient magnitude of the distillation loss $\mathcal{L}_{\mathrm{Distill.}}$ is much larger than that of the reconstruction loss $\mathcal{L}_{\mathrm{reconst.}}$ during the early training phase. Starting with a chaotic urban 3D representation under the influence of the distillation loss caused the reconstruction loss to struggle to maintain consistency between the repaired regions and the observable scene distribution, ultimately leading to training failure. ii) The Gaussian parameters used in the second stage were systematically determined through controlled experiments. We evaluated the performance under various parameter settings, and the current configuration has demonstrated the best results.

\subsection{Training Time}

Similar to VEGS, VSD-based diffusion knowledge distillation is widely known to be slow. To address this, we employed the SDS-based method to improve speed, reducing the process to about two iterations per second. EVS setting generates more extra content. UrbanCraft and VEGS must maintain consistency with the reconstructed content, which increases the learning difficulty and optimization time.

\subsection{Loss Explanation}

HSG-VSD leverages hierarchical sem-geometric priors into proposed HSG-based score distillation to ensure 3D consistency with reconstruction during optimization. Geometry SDS refines depth and normal maps using normal score distillation to enforce realistic geometry and reduce artifacts in extrapolated views.

\subsection{Performance divergence}

VEGS acknowledged that their method struggles with unobserved spaces generated by large-angle EVS, which impacts visualization results. Benchmark~\cite{han2024extrapolatedurbanviewsynthesis} also highlighted this issue, leading VEGS to \textbf{remove half of the image plane for improved visualization}. In contrast, in the supplementary section, we tackle the challenging task of repairing large-angle invisible synthesis, an issue left unresolved by VEGS and largely overlooked by the current SoTA in the EVS field. To highlight the current state of wide-angle EVS, we present the most original rendering results \textbf{under large-angle EVS setting} in supplementary materials.

%

\section{Ablation Study}

\subsection{Geometry Consistency} 
In Figure~\ref{fig:supp3}, we illustrate the visualization results of depth maps and RGB images. It is evident that our method significantly outperforms existing approaches in reconstructing geometric structures and preserving texture details. In particular, our method excels in complex areas, such as object edges and fine textures, thanks to the proposed UrbanCraft. Moreover, our approach effectively reduces noise and artifacts, delivering depth maps with superior consistency and clarity. Meanwhile, Figure~\ref{fig:supp5} further presents the comparison results of depth maps and RGB images, verifying the effectiveness of the proposed $\mathcal{L}_{\mathrm{G-SDS}}$ in our method. We note that: i) the proposed $\mathcal{L}_{\mathrm{G-SDS}}$ significantly diminishes surface cavities on vehicles and mitigates the sharp cone artifacts, resulting in notably smoother vehicle geometries, and ii) the original Gaussian optimization process can introduce floating artifacts due to a lack of sufficient regularization. To overcome this, the proposed $\mathcal{L}_{\mathrm{G-SDS}}$ incorporates rich priors from the large-scale text-to-image diffusion model, providing enhanced regularization for the associated geometry distribution. Hence, by employing the full method, we achieve the most consistent and clear depth maps and RGB images, further demonstrating the effectiveness and superiority of the proposed approach.

\subsection{HSG-SDS} We also validate the effectiveness of the proposed HSD-VSD compared to the HSG-SDS variant. Specifically, original SDS achieves text-to-3D generation by distilling a pre-trained text-to-image diffusion model $\epsilon_{p}$ to optimize a differentiable 3D representation parameterized by $\theta$. Following HSG-VSD, the HSG-SDS variant utilizes the 2D control signals $\mathcal{C}$ from the sem-geometric representations $SG$ at the extrapolated camera view $\mathbf{T}$. Subsequently, the features produced by ControlNet $\psi(\cdot)$ are integrated into the diffusion model $\epsilon_{p}$. Finally, the gradient of HSG-SDS loss can be formulated as:
\begin{equation}
\resizebox{0.70\hsize}{!}{$
\begin{split}
    \label{eq:self}
    \nabla_{\theta}\mathcal{L}_{\mathrm{HSG-VSD}}(\theta) & \triangleq \mathbb{E}_{t, \epsilon, \mathbf{T}}[\omega(t)(\epsilon_{p}(\boldsymbol{x}_{t},t,y,\\& \psi(SG(\mathbf{T}))-\epsilon)\frac{\partial \boldsymbol{g}(\theta, \mathbf{T})}{\partial \theta}],
\end{split}$}
\end{equation}
where random noise $\boldsymbol{\epsilon} \sim \mathcal{N}(\boldsymbol{0}, \boldsymbol{\textit{I}})$, $t \sim \mathcal{U}(0.02,0.98)$, $\omega(t)$ weights the loss given the time step $t$, and $\boldsymbol{x}_{t}$ is the randomly perturbed rendered image. Although SDS can produce reasonable 3D content aligned with the text prompt, it exhibits over-saturation, over-smoothing, and low diversity.

\section{Discussion}

\subsection{Challenge Task: Barely-visible Vehicle}

\begin{figure}[t]
    \centering
    \includegraphics[width=\linewidth]{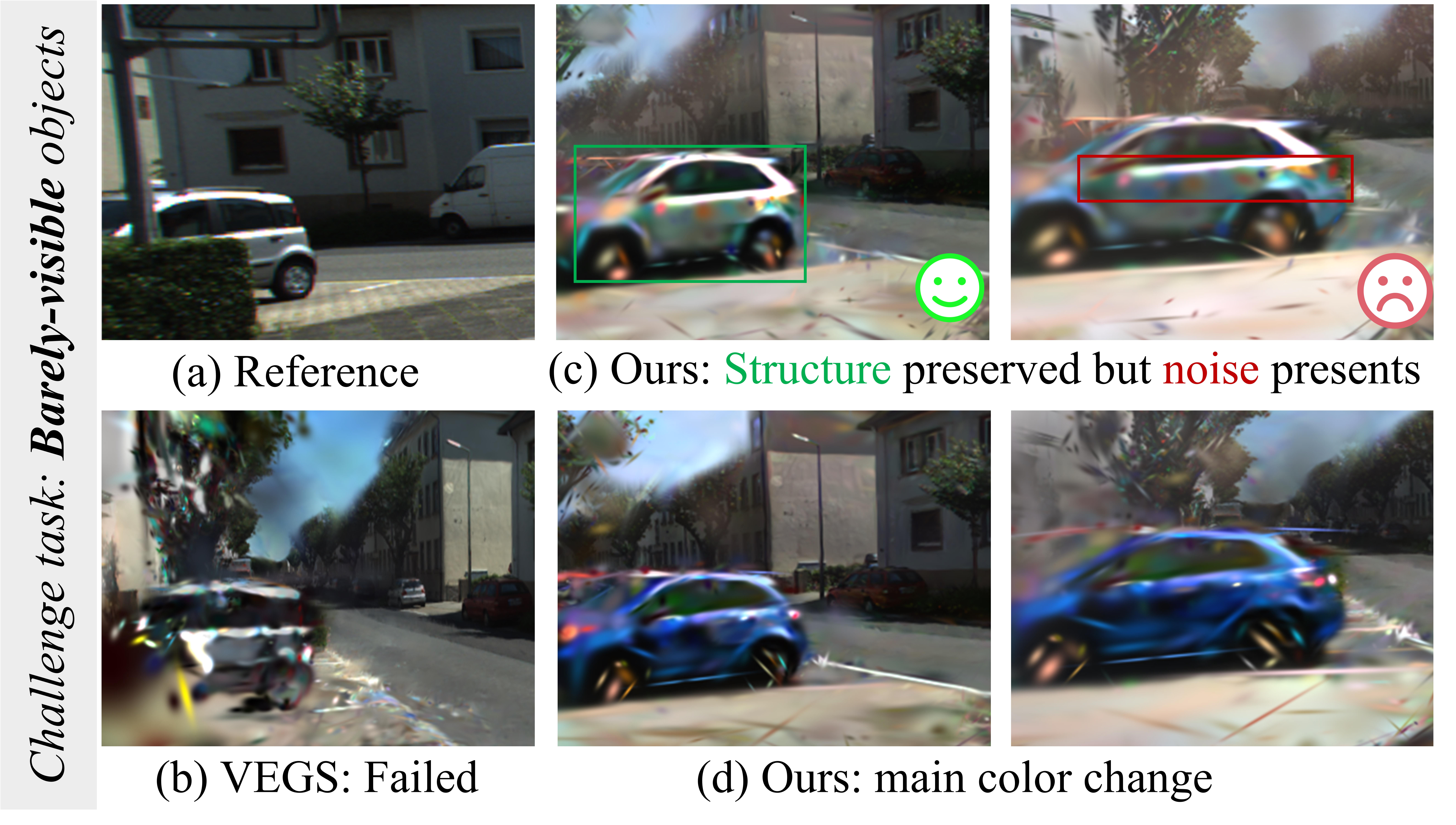}
    \vspace{-5mm}
    \caption{Comparison of different methods on the barely-visible objects challenge. (a) Reference image with a barely visible vehicle. (b) VEGS fails to reconstruct the vehicle structure, leading to severe distortions. (c) Our method accurately restores the vehicle’s structure with clear part attributes, though slight noise is present. (d) Our method also enables color editing, successfully transforming the silver car into a blue one.}
    \label{fig:supple_chall}
    \vspace{-3mm}
\end{figure}

Figure~\ref{fig:supple_chall} showcases our method's effectiveness in handling barely visible objects, particularly for vehicle restoration. Unlike VEGS, which fails to reconstruct the car structure, our approach accurately preserves the vehicle’s geometric integrity and part details. Even under challenging visibility conditions, our method ensures that the car’s shape and attributes remain recognizable. Furthermore, our method enables appearance modification by leveraging UrbanCraft-guided sampling. As demonstrated in Figure~\ref{fig:supple_chall}.d, the original silver car is successfully transformed into a blue one while maintaining its structural fidelity.

\subsection{Failure Cases Analysis and Limitation}

We have observed the following failure cases and limitations in our method: i) \textbf{Gaussian Sparks in Unseen Extrapolated Views}. As shown in Figure~\ref{fig:supp7}.\textcolor[rgb]{0.21,0.49,0.74}{(a,b,i)}, the sparks tend to occur in large unseen extrapolated camera views. Theoretically, the gradients from HSG-VSD guidance during backpropagation are relatively large~\cite{poole2022dreamfusion, wang2024prolificdreamer}. Hence, to balance the reconstruction and HSG-VSD gradients, we utilize the loss weights  $\lambda_{\mathrm{reconst}}=1e4, \lambda_{\mathrm{HSG-VSD}} = 1.0$ to achieve it. However, in these challenging views, the reconstruction loss is particularly sparse, resulting in optimization imbalance. Consequently, the gradients from the HSG-VSD guidance loss significantly exceed those from the reconstruction loss, leading to the appearance of the sparks, and ii) \textbf{Blurring Artifacts in Incomplete Occupancy Grids}. As shown in Figure~\ref{fig:supp7}.\textcolor[rgb]{0.21,0.49,0.74}{(c,d,e,f)}, blurring artifacts are evident. These issues primarily arise when the occupancy grid information for the scene is incomplete or when the observed regions extend beyond the perceptual range of the occupancy grid. Under such conditions, the proposed HSG-VSD struggles to distill effective information from the pretrained UrbanCraft2D model, resulting in suboptimal optimization of the urban 3D representation and the emergence of blurring artifacts. We hope our method serves as a starting point for urban view extrapolation tasks, leaving the resolution of these challenges for future work. 

\subsection{Compare with 3D Generation Baselines}

\begin{figure}[t]
    \centering
    \includegraphics[width=\linewidth]{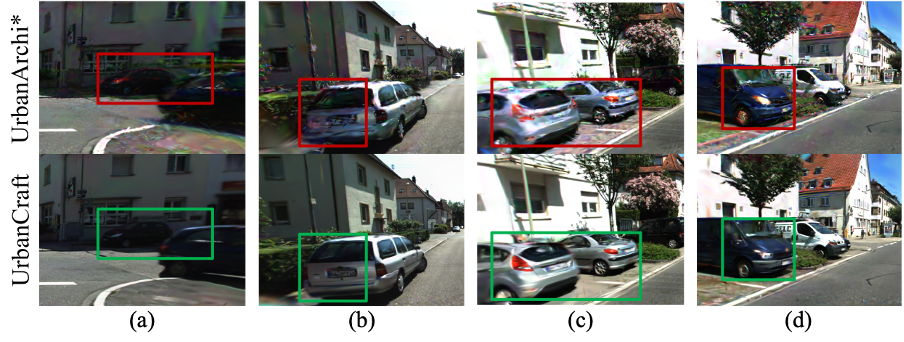}
    \vspace{-4ex}
    \caption{Qualitative comparison between UrbanArchi* and proposed UrbanCraft.}
    \label{fig:supp_archi}
    \vspace{-2mm}
\end{figure}

\textbf{Differences in problem setting:} Compared to existing generation tasks, wide-angle EVS requires finer-grained control capabilities.  This not only requires the ability to generate new content but also ensures that the generated parts maintain a high degree of geometric and semantic consistency with the existing reconstructed content. ii) \textbf{Limitations of existing generation tasks:} if existing generation tasks are directly applied to EVS, as shown in Figure~\ref{fig:supp_archi}. To ensure fairness, we replaced the original hash grid representation of Urban Architect with 3D Gaussians and named it UrbanArchi*. From Figure~\ref{fig:supp_archi}, the UrbanArchi* has incorrect geometric-semantic modifications to the foreground of EVS, while our proposed UrbanCraft can effectively address this issue by proposed hierarchical sem-geometric priors, particularly layout-based pose control, enabling effective control in EVS generation.

\section{UrbanCraft on NuScenes Dataset} 

\begin{figure}[t]
    \centering
    \includegraphics[width=0.6\linewidth]{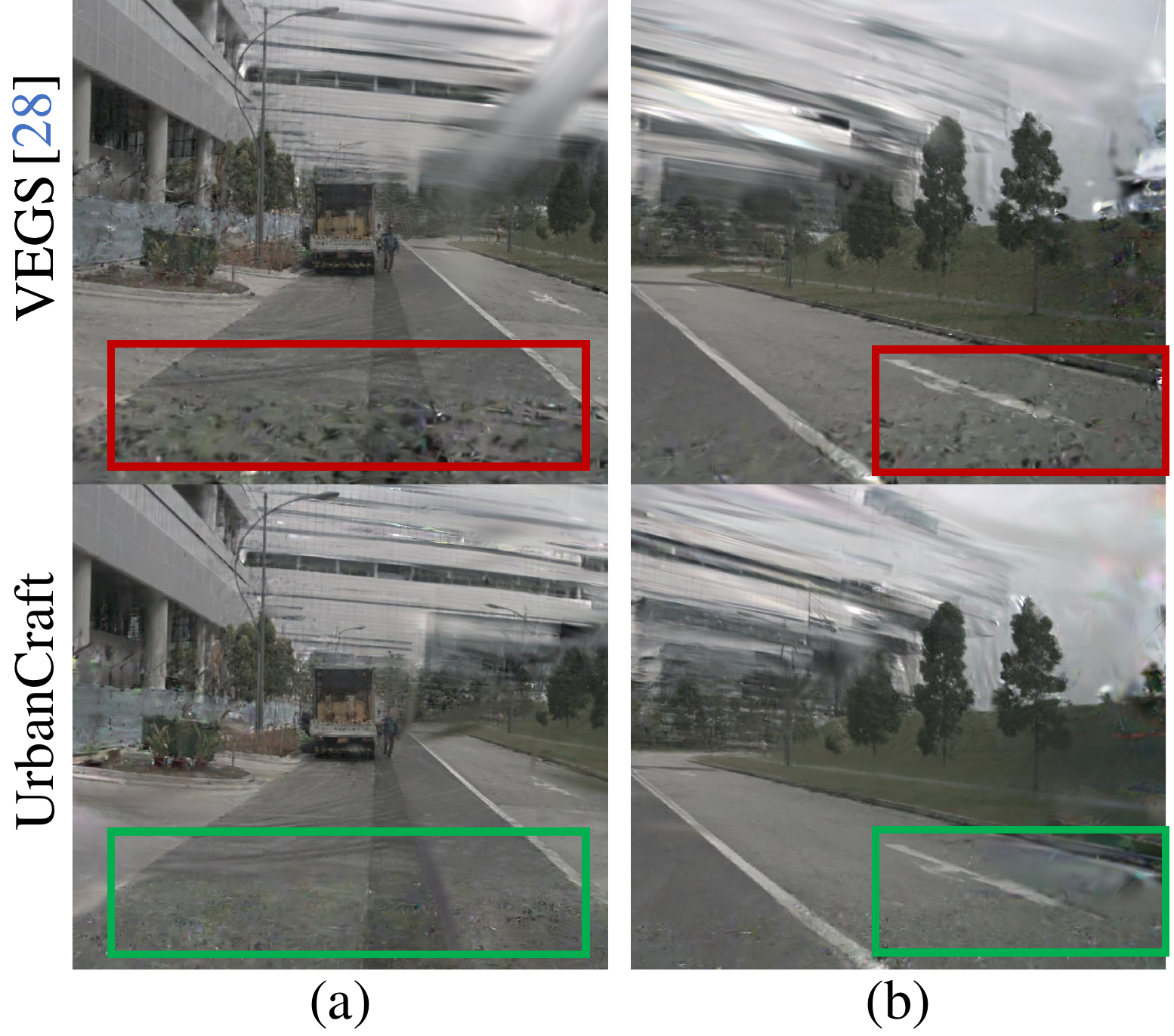}
    \vspace{-2mm}
    \caption{Qualitative comparison on NuScenes~\cite{caesar2020nuscenes} for extrapolated view synthesis.}
    \label{fig:supp8}
    \vspace{-2mm}
\end{figure}

In the main manuscript, to ensure a fair comparison, we utilize the same KITTI-360~\cite{liao2022kitti} dataset as VEGS~\cite{hwang2024vegs} for both qualitative and quantitative analysis. To further validate the effectiveness of the proposed UrbanCraft, we developed a dedicated DataLoader for the nuScenes~\cite{caesar2020nuscenes} dataset and applied it to all baseline methods. Specifically, for a given training sequence, we stack the key frame's point cloud information as the initialization for 3D Gaussians~\cite{kerbl20233d}, using the same hyper-parameter settings as those for the KITTI-360 dataset. The detailed results are presented in Figure~\ref{fig:supp8}. 

We can see that: i) the proposed UrbanCraft demonstrates its capability to address similar challenges on the nuScenes dataset. For instance, it effectively resolves issues like uneven ground surfaces and extrapolated camera views with significant occlusions and text-ambiguous descriptions, significantly improving the overall smoothness and coherence of the RGB images.

\clearpage
\section{Public Resources Used}

In this section, we acknowledge the public resources used, during the course of this work.

\subsection{Public Datasets Used}
\begin{itemize}
    \item KITTI-360\footnote{\url{https://www.cvlibs.net/datasets/kitti-360}} \dotfill CC BY-NC-SA 3.0
    \item NuScenes\footnote{\url{https://www.nuscenes.org}} \dotfill Apache License 2.0
    \item SSCBench\footnote{\url{https://github.com/ai4ce/SSCBench}} \dotfill CC BY-NC-SA 4.0
    \item Occ3D\footnote{\url{https://github.com/Tsinghua-MARS-Lab/Occ3D}} \dotfill MIT License
\end{itemize}

\subsection{Public Implementations Used}
\begin{itemize}
    \item Urban Architect\footnote{\url{https://github.com/UrbanArchitect}} \dotfill MIT License
    \item VEGS\footnote{\url{https://github.com/deepshwang/vegs}} \dotfill Software License
    \item NuScenes-devkit\footnote{\url{https://github.com/nutonomy/nuscenes-devkit}} \dotfill Apache License 2.0
    \item NSVF\footnote{\url{https://github.com/facebookresearch/NSVF}} \dotfill MIT License
    \item Semantic-kitti-api\footnote{\url{https://github.com/PRBonn/semantic-kitti-api}} \dotfill MIT License
\end{itemize}

\end{document}